\documentclass[final]{cvpr}

\usepackage{times}
\usepackage{epsfig}
\usepackage{graphicx}
\usepackage{amsmath}
\usepackage{amssymb}
\usepackage{textcomp}
\usepackage{xr-hyper}

\usepackage{amsmath,amsfonts,bm}

\def\eqref#1{equation~\ref{#1}}

\def\1{\bm{1}}

\def\ry{{\textnormal{y}}}

\def\rmZ{{\mathbf{Z}}}

\def\vmu{{\bm{\mu}}}

\def\vw{{\bm{w}}}

\def\vy{{\bm{y}}}

\def\evw{{w}}

\def\evy{{y}}

\def\mI{{\bm{I}}}

\def\mZ{{\bm{Z}}}

\def\mSigma{{\bm{\Sigma}}}

\DeclareMathAlphabet{\mathsfit}{\encodingdefault}{\sfdefault}{m}{sl}
\SetMathAlphabet{\mathsfit}{bold}{\encodingdefault}{\sfdefault}{bx}{n}

\newcommand{\E}{\mathbb{E}}

\newcommand{\R}{\mathbb{R}}

\newcommand{\KL}{D_{\mathrm{KL}}}

\usepackage[pagebackref=true,breaklinks=true,colorlinks,bookmarks=false]{hyperref}

\begin{document}

\title{Enriching ImageNet with Human Similarity Judgments and Psychological Embeddings}

\author{Brett D.~Roads \& Bradley C.~Love \\
Department of Experimental Psychology \\
University College London \\
London, United Kingdom \\
{\tt\small \{b.roads,b.love\}@ucl.ac.uk} \\
}

\maketitle

\begin{abstract}
Advances in object recognition flourished in part because of the availability of high-quality datasets and associated benchmarks. However, these benchmarks---such as ILSVRC---are relatively \emph{task-specific}, focusing predominately on predicting class labels. We introduce a publicly-available dataset that embodies the \emph{task-general} capabilities of human perception and reasoning. The Human Similarity Judgments extension to ImageNet (ImageNet-HSJ) is composed of human similarity judgments that supplement the ILSVRC validation set. The new dataset supports a range of task and performance metrics, including the evaluation of unsupervised learning algorithms. We demonstrate two methods of assessment: using the similarity judgments directly and using a \emph{psychological embedding} trained on the similarity judgments. This embedding space contains an order of magnitude more points (i.e., images) than previous efforts based on human judgments. Scaling to the full 50,000 image set was made possible through a selective sampling process that used variational Bayesian inference and model ensembles to sample aspects of the embedding space that were most uncertain. This methodological innovation not only enables scaling, but should also improve the quality of solutions by focusing sampling where it is needed. To demonstrate the utility of ImageNet-HSJ, we used the similarity ratings and the embedding space to evaluate how well several popular models conform to human similarity judgments. One finding is that more complex models that perform better on task-specific benchmarks do not better conform to human semantic judgments. In addition to the human similarity judgments, pre-trained psychological embeddings and code for inferring variational embeddings are made publicly available. Collectively, ImageNet-HSJ assets support the appraisal of internal representations and the development of more human-like models.
\end{abstract}

\section{Introduction}
\label{intro}

One interesting question is how models' internal representations compare to human-perceived similarities.  While people make such judgments with little effort, human-perceived similarity flexibly adapts to different contexts, reflecting a rich understanding of the world \cite{MurphyMedin1985:289,Jones_Love_2007}. For example, people may perceive a beer bottle as similar to cigarettes because both are age-restricted, while perceiving a beer bottle as similar to a soda because both are beverages. Humans may perceive two objects as similar for many reasons, including the two objects playing related roles within encompassing systems, sharing perceptual properties, or simply interacting with one another \cite{Jones_Love_2007}. Human-perceived similarity has been leveraged in applications such as image retrieval \cite{ElNaga_etal_2004,Guo_etal_2002} and human-in-the-loop categorization \cite{Roads_Mozer_2017,WahEtal2015,Zhen_etal_2006}, but has been under-utilized in the general development of computer vision algorithms. The lack of research is partly due to the absence of an appropriate dataset and the technical challenges associated with collecting such a dataset. This work introduces the Human Similarity Judgments extension to ImageNet (ImageNet-HSJ), designed to include maximally informative similarity judgments for the widely-used ILSVRC dataset \cite{Deng_etal_2009, ILSVRC15}.

As algorithm development shifts from learning \emph{task-specific} representations towards \emph{task-general} representations, the evaluation metrics used to assess models may also benefit from an equivalent shift. While task-specific metrics---such as classification accuracy---will always be relevant, complementary task-general evaluation metrics seem increasingly necessary. One strategy for creating a task-general metric is to assess how well model-perceived similarity aligns with human-perceived similarity. Comparing human and model similarity allows researchers to focus on the internal representations that precede task-specific output. Internal representation metrics create a level playing field when comparing across diverse training paradigms; such as supervised, unsupervised, and self-supervised approaches. Helping machines think in a more human-like way may also improve human-machine interactions.

The main aims of this work are to assemble a dataset that embodies human-perceived similarity and demonstrate how the dataset can be used to assess arbitrary models. Extending previous work \cite{Tamuz_2011_adaptively, Roads_Mozer_2019_BRM}, we employed \emph{psychological embeddings} to concisely model the information contained in the similarity judgments. A psychological embedding includes an embedding of the stimuli, as well as functions that link the embedding to observed behavior. The psychological embeddings serve three different roles: as a means of modeling uncertainty and performing active learning (\S~\ref{sec:collection}), assessing whether sufficient data has been collected (\S~\ref{sec:convergence}), and evaluating the internal representations of arbitrary models (\S~\ref{sec:evaluation}). To handle the large number of stimuli, existing approaches are extended using variational inference and ensembles. The dataset and companion pre-trained psychological embeddings are hosted at \url{https://osf.io/cn2s3/}. An open-source python package for inferring variational psychological embeddings is available at \url{https://github.com/roads/psiz}.

\section{Related work}
\label{sec:related}

There is a long history of using human similarity judgments to infer embeddings \cite{Torgerson_1952,Torgerson_1958}. The family of algorithms includes a number of kernel variants \cite{Gower_1966,Nosofsky_1985:415,Shepard_1957,Tamuz_2011_adaptively,vanderMaaten2012:MLSP1}, as well as non-metric approaches \cite{Shepard1962:125,shepard1962:219}. More recent work has seen an increase in the scale and naturalism of the stimuli being used  \cite{Wilber_2014_cost,Wah_2014,Roads_Mozer_2019_BRM,Roads_etal_2018,Nosofsky_2018_rocks}. Of particular note is recent work that collected similarity judgments for 1,854 unique images belonging to the THINGS dataset \cite{Hebart_2019_THINGS,Hebart_Zheng_Pereira_Baker_2020}. In this work, we advance the size of the stimulus set by an order of magnitude, while also focusing on a dataset that is widely used in machine learning.

An enormous volume of work has conducted targeted comparisons between artificial neural networks and human behavior. Much of this work has focused on comparing human and model \emph{classification} performance \cite{Golan_etal_2020_controversial, Peterson_etal_2019_ICCV}. Classification experiments have revealed differences in image-level confusion statistics \cite{Rajalingham_etal_2018}, as well as the degree of shape bias \cite{Baker_etal_2018_deep}. Other work has examined the ability of artificial neural networks to predict human-perceived typicality ratings \cite{Lake_etal_2015}. Most relevant is work that compared human- and model-perceived \emph{similarity}. For example, a VGG16 \cite{Simonyan_Zisserman_2015} model pretrained on ImageNet can predict both human similarity ratings \cite{Peterson_Abbot_Griffiths_2018} and similarity rankings \cite{Attarian_etal_2020} at above chance levels. Taking things even further, Attarian \etal, Peterson \etal, and Sanders \etal \cite{Sanders_2020_dnn} have demonstrated that neural networks can be specifically trained to predict human similarity judgments with relatively high fidelity. Our aim is to complement existing research by providing a dataset that can serve many purposes, in addition to being an evaluation metric.

A number of active learning paradigms exist for efficiently collecting data \cite{Jamieson_etal_2015, Rau_etal_2016, Sievert_etal_2017}. This work builds on an approach specifically catered to collecting human similarity judgments \cite{Tamuz_2011_adaptively, Roads_Mozer_2019_BRM}. While Roads \etal \cite{Roads_Mozer_2019_BRM} used MCMC to obtain posteior samples, we directly sample from an approximate posterior learned using variation inference. This approach reduces computational costs, helping us scale up to larger stimulus sets. In addition to computational savings, variational inference also eliminates a problem associated with drawing posterior samples from an embedding. Embeddings typically have some degree of invariance, such as being rotation invariant in Euclidean space. Consequently, sampling the posterior using MCMC can artificially inflate the uncertainty of peripheral points, as the sampling chain rotates in space. While various techniques can be used to limit this issue \cite{Gronau_Lee_2020}, variational inference naturally addresses the problem.

\section{ImageNet-HSJ dataset}

The Human Similarity Judgments extension to ImageNet (ImageNet-HSJ) is a versioned dataset composed of ordinal human similarity judgments intended to supplement the ILSVRC 2012 validation dataset. This work focuses exclusively on the validation set because evaluation is a prerequisite for model selection paradigms. The dataset includes a large number of quality-controlled similarity judgments, collected from a wide range of participants (Table~\ref{tab:dataset}).

ImageNet is an ideal dataset for enrichment because it is widely used and captures a spectrum of everyday visual experience. In addition to exhibiting both conceptual breadth and depth, each class is made up of highly variable images. ImageNet also strikes a balance in size. While small relative to more recent datasets, it is still challenging enough to remain relevant. The remainder of this section describes the human task, stimuli coverage, and the versioning strategy.

\begin{table}
\begin{center}
\begin{tabular}{|l|c|c|}
\hline
Property & Seed (v0.1) & Full (v0.2) \\
\hline\hline
Unique stimuli & 1,000 & 50,000 \\
Judged trials & 25,273 & 384,277 \\
Unique participants & 431 & 5572 \\
Median trial duration (s) & 7.85 & 8.30 \\
Participant contribution & & \\
\hspace{3mm}Median & 0.18\% & 0.01\% \\
\hspace{3mm}Max & 3.48\% & 1.16\% \\
\hline
\end{tabular}
\end{center}
\caption{General properties of ImageNet-HSJ broken down by version.}
\label{tab:dataset}
\end{table}

\subsection{Human similarity judgment task}

Human similarity judgments were collected using a web-based application. Participants were recruited via Amazon Mechanical Turk (AMT) and completed a \emph{session} in exchange for monetary compensation (approximately 8.00 USD per hour). Each session was composed of 50 trials and designed to take 10 minutes.

\begin{figure}[t]
\label{fig:trial_format}
\begin{center}
\includegraphics[width=1.\linewidth]{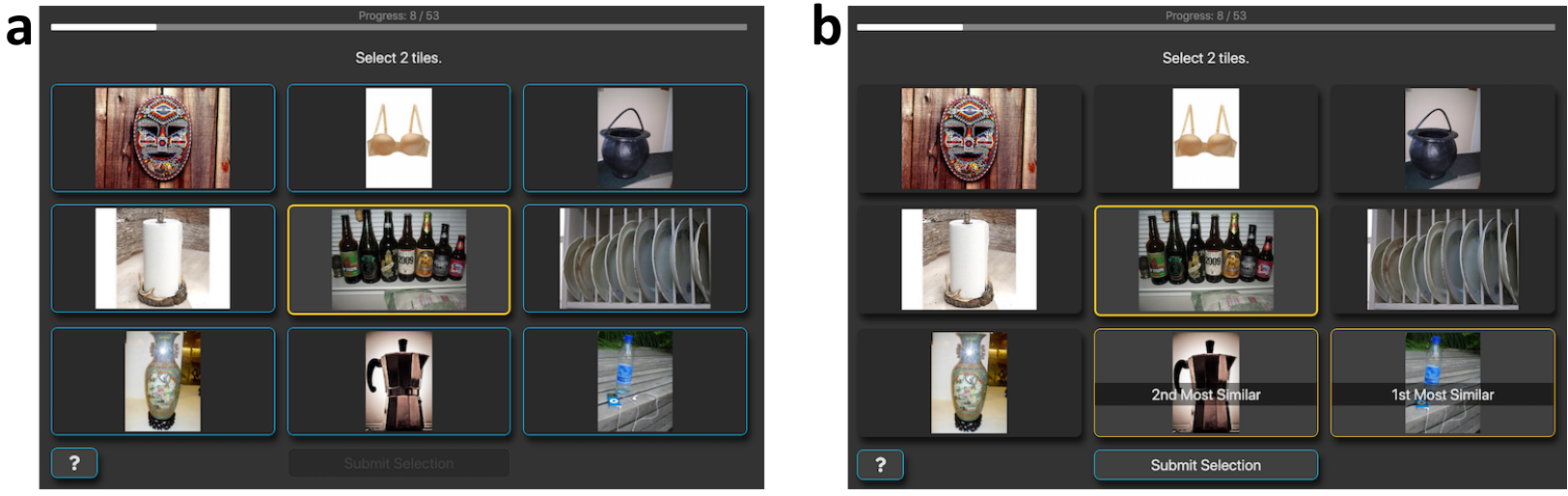}
\end{center}
\caption{An example similarity judgment trial shown to human participants. \textbf{(a)} On initial presentation, nice images are shown with the query image placed in the center. \textbf{(b)} When participants click a tile, it is highlighted and marked as their first or second choice. After participants select two reference images, they are allowed to advance to the next trial. In this case, the participant thought the beer bottles were most similar to a plastic water bottle, and second most similar to a moka pot.}
\end{figure}

On each trial, participants were presented with a display composed of nine images arranged in a grid (Fig~\ref{fig:trial_format}). The center image is the \emph{query} and the surrounding images are \emph{references}. Participants were instructed to select the two reference images most similar to the query. Based on the order of their selection, participants also indicated which reference they thought was most similar and second most similar. Participants could unselect a reference by clicking it a second time. Once satisfied, participants clicked a button to submit the trial and the next trial was displayed. Participants could re-view the instructions at any time by clicking a question mark button.

In general, a trial can be composed $R$ references and participants tasked with making $R-1$ choices \cite{Wah_2014, Wilber_2014_cost, Roads_Mozer_2019_BRM}. The 8-rank-2 trial format was chosen because most participants find it an easy task and it strikes a reasonable balance between throughput and noise \cite{Demiralp_etal_2014, Li_etal_2016}.

\subsection{Stimulus set}

Human similarity judgments were collected for all 50,000 images of the ILSVRC validation set. From the \emph{full} dataset, we selected a subset of stimuli to serve as the \emph{seed} subset. Data collection began by focusing exclusively on collecting similarity judgments for the seed subset, helping the active learning procedure get off the ground. Once convergence was achieved (see \S~\ref{sec:convergence}), collection began for the full dataset.

The seed subset is composed of 1,000 images with a single representative image for each class. The representative images were selected by using a pre-trained VGG19 network to make predictions for all images in the ILSVRC validation set. For each class, the image with the highest probability of correct classification was chosen as the representative. In the case of ties, an image was randomly selected from the top performers. Images that were added to the exclusion list in ILSVRC 2014 were not eligible to be part of the seed subset.

\subsection{Versioning}

ImageNet-HSJ releases follow a versioning scheme. Version 0.1 includes similarity judgments from the initial collection phase that only included the seed subset. Version 0.2 expands coverage to the full stimulus set and includes all judgments collected at the time of submission. Once convergence is achieved for the full dataset (see \S~\ref{sec:convergence}), a final version 1.0 will be released.

\section{Dataset collection}
\label{sec:collection}

The primary goal of the data collection process is to assemble a set of observations that informatively probes human-perceived similarity. Given a stimulus set of 50,000 images and an 8-rank-2 trial format, there are more than $10^{36}$ potential trials to show participants. Given the conceptual breadth of ImageNet, most trials are likely to be composed of images that are all highly dissimilar (\eg, firetruck, banana, squid, baseball, parrot, elephant, tiara, ladybug, and pretzel). On these trials participants are likely to make highly idiosyncratic or random choices. Instead, we would like to prioritize trials that are reveal consistent perceptual beliefs.

\begin{figure}[ht]
\label{fig:active_learning}
\begin{center}
\includegraphics[width=1.\linewidth]{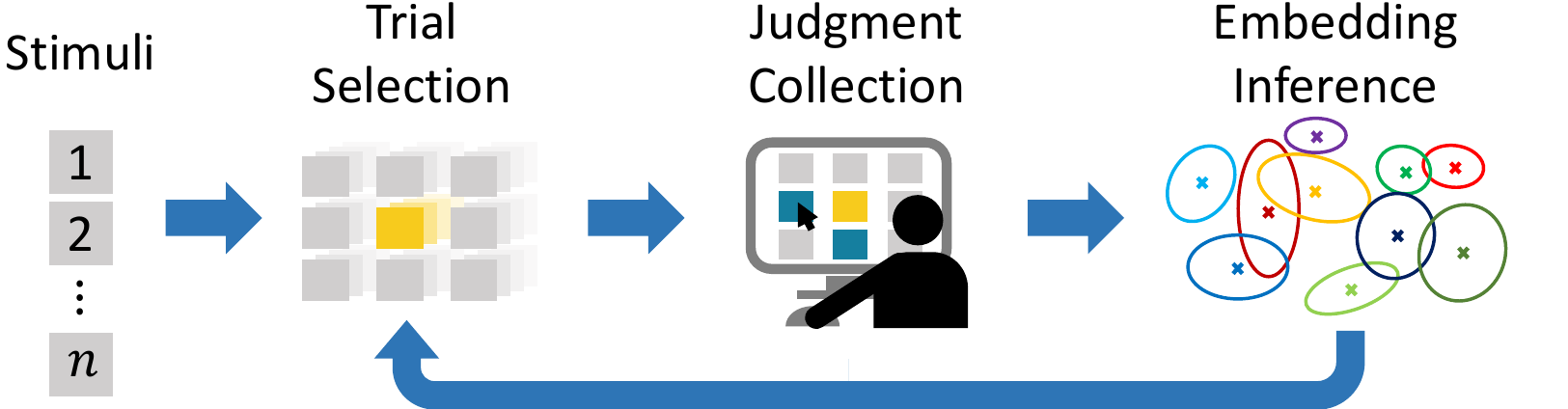}
\end{center}
\caption{Outline of the active learning procedure. A set of trials is selected from a set of stimuli. Following human judgment, variational inference is used to obtain posterior distributions for an ensemble of psychological embeddings. The posterior distributions are used to select the next set of trials that maximizes expected information gain. After each iteration, the increase in information will result in increasingly precise estimates of where stimuli reside in the embedding.}
\end{figure}

In lieu of random sampling, we use an active learning paradigm to identify promising trials across multiple iterations. We extend existing active learning paradigms \cite{Sievert_etal_2017, Tamuz_2011_adaptively, Roads_Mozer_2019_BRM} using variational Bayesian inference. Each iteration of the active learning procedure is composed of three steps: trial selection, judgment collection and embedding inference (Figure~\ref{fig:active_learning}). In the first step, an existing psychological embedding is used to select a new set of trials that maximizes expected information gain (\S~\ref{sec:trial_selection}). In the absence of data, the psychological embedding is governed by a prior distribution and trials are selected randomly. In the second step, human similarity judgments are collected (\S~\ref{sec:judgment_collection}). In the third step, all available observations are used to infer a new ensemble of psychological embeddings (\S~\ref{sec:embedding_inference}).

\subsection{Judgment collection}
\label{sec:judgment_collection}

Trials selected by the active learning procedure were randomly partitioned into 72 sessions composed of 50 trials each. During a 50-trial session, there were four \emph{catch trials} designed to assess quality and earnest effort. Catch trials were identical to randomly sampled trials, except that one of the reference images was a mirror-image of the query image. Catch trials were strategically placed such that two catch trials occured in the first 20 trials and two in the last 20 trials.

For a given session, catch trial performance determined the sample weight of the trials. For each catch trial, a participant can either select the mirrored query as the first choice, second choice, or not at all; resulting in a grade of 1, .5, and 0 respectively. The sample weight of each trial is defined as the mean grade of all the catch trials.

Depending on the session's average grade, a session was marked \emph{premium} ($>.875$), \emph{satisfactory} ($[.5, .875[$), or \emph{unsatisfactory} ($<.5$). Unsatisfactory sessions were dropped from the final dataset. If a participant received a premium grade, they remained eligible to complete more sessions. If a participant received a non-premium grade, they were added to an ineligible list. A scheduler created HITs on AMT in order to obtain one premium rating for each session.

As an additional precaution, we also dropped any trials with a duration less than 1 s, which is extremely rare. We did not require trials be under a specific duration since there are many valid reasons for a long trial that should not impact quality.

\subsection{Embedding inference}
\label{sec:embedding_inference}

Given a training dataset composed of trials, observed outcomes, and sample weights $\mathcal{D}=\{ \mathcal{T}, \vy, \vw \}$, our goal is to infer an ensemble of psychological embeddings (Fig~\ref{fig:active_learning}). The remainder of this section details inference for a single embedding $\mZ \in \R^{n\times d}$, where $n$ indicates the number of stimuli and $d$ is a hyperparameter that determines the dimensionality of the embedding. For improved readability, $\mZ_{i}$ is used to denote the $i$th row vector (\ie, the $i$th stimulus embedding).

The set of $m$ trials is defined as $\mathcal{T} = \{\mathcal{T}_{1}, \ldots, \mathcal{T}_{m}\}$. For each trial, $\mathcal{T}_{i} = \{q_{i}, \mathcal{R}_{i}\}$, where $q_{i}$ indicates the query stimulus index and $\mathcal{R}_{i}$ indicates an ordered $n$-tuple of reference stimuli indices. The references are ordered so that a deterministic rule can be applied to enumerate all possible outcomes associated with a particular trial. In general, a trial with $r$ references requiring $c$ ranked choices has $k = \frac{r!}{\left(r - c\right)!}$ distinct outcomes. Thus $\evy_{i}$ indicates the categorical outcome of the $i$th trial and $\evw_{i}$ the corresponding sample weight. Sample weights are determined by a trial's corresponding session grade.

Assuming independence between trials, the likelihood function is
\begin{equation}
p\left( \mathcal{D} | \mZ \right) =  \prod_{i}^{m} p\left( \evy_{i} | \mathcal{T}_{i}, \mZ \right)^{\evw_i}.
\end{equation}
Sample weights are integrated into the likelihood so that the weights linearly scale each trial's contribution in logarithmic space. Following \cite{Roads_Mozer_2019_BRM}, the embedding is linked to observed behavior using a Luce's ratio of strengths formulation \cite{Luce1959} that is widely used in psychology \cite{Shepard_1957, Shepard_1958, Nosofsky_1985:415, Nosofsky_1986}. For a single 8-rank-2 trial, the likelihood is
\begin{equation}
p\left( \evy_{i} | \mathcal{T}_{i}, \mZ \right) = \frac{s\left(\mZ_{q_i}, \mZ_{a} \right)}{\sum_{r \in \mathcal{R}_{i}} s\left(\mZ_{q_i}, \mZ_{r} \right)}\frac{s\left(\mZ_{q_i}, \mZ_{b} \right)}{\sum_{r \in \mathcal{R}_{i}\neg a} s\left(\mZ_{q_i}, \mZ_{r} \right)}.
\end{equation}
The subscripts $a$ and $b$ are semantic pointers to reference indices in $\mathcal{R}_{i}$. The subscript $a$ indicates the first choice reference and $b$ indicates the second choice reference. Motivated by psychological theory \cite{Shepard_1957, Shepard_1958, Nosofsky_1985:415, Nosofsky_1986}, this work assumes the similarity function
\begin{equation}
s\left(\mZ_{i}, \mZ_{j} \right) = \exp\left(-\beta || \mZ_{i} - \mZ_{j} || \right),
\end{equation}
although alternative functions could also be used \cite{Agarwal_etal_2007, Tamuz_2011_adaptively, vanderMaaten2012:MLSP1}.
In this work, $\beta$ serves as a convenience parameter and is set to $10$. The prior distribution of the embedding is
\begin{equation}
p(\mZ_{i}) = \mathcal{N} ( 0, \sigma \mI),
\end{equation}
where $\sigma$ is a scaling parameter fit with the other free parameters.

To perform variational inference, the true posterior $p(\mZ | \mathcal{D})$ is approximated with the variational distribution $q(\mZ | \mathcal{D})$ that assumes
\begin{equation}
\rmZ_{i} \sim \mathcal{N} \left( \vmu^{(i)}, \mSigma^{(i)} \right).
\end{equation}
To limit model complexity, $\mSigma^{(i)}$ is constrained to be a diagonal covariance matrix. The set of all free parameters is denoted by $\Theta = \{\vmu^{(i)}, \mSigma^{(i)}, \sigma | 1 \le i \le n \}$. The optimization objective is to minimize the variational free energy
\begin{equation}
L(\mathcal{D}, \Theta) = \KL (q(\rmZ|\Theta ) \Vert p(\rmZ)) - \E_{q(\rmZ | \Theta)} \log p(\mathcal{D}|\rmZ),
\end{equation}
where the first term is the complexity cost and the second term is the likelihood cost \cite{Hoffman_etal_2013, Ranganath_etal_2014}. After minimizing the variational free energy, the approximate posterior distribution provides a concise description of the uncertainty associated with each embedding point. It is worth noting that posterior uncertainty includes both epistemic and aleatoric uncertainty.

Embedding algorithms are sensitive to initial conditions and prone to discovering local optima. In a low-data regime, different weight initializations can result in solutions that provide equally good fits to the training data, but exhibit qualitatively different structure. Since the embeddings are used downstream to select future trials, we limit the influence of any single model by using an ensemble composed of three equally weighted embeddings. Similar to standard bootstrap aggregating, variability is encouraged by withholding a different 5\% subset of the training data for each embedding.

The 5\% withheld data determined the dimensionality hyperparameter $d$. For a given iteration, independent ensembles were inferred with different dimensionality values. The ensemble with the best categorical cross entropy determined the selected dimensionality. Since inferring models is somewhat costly, this process was not performed at every iteration. Instead, the dimensionality was set based on the chosen dimensionality of the previous iteration.

At the beginning of each iteration, the ensemble was partially initialized using weights from the previous iteration. Using categorical cross-entropy on the validation set, the two top performing models resumed training from the weights of the previous iteration. The worst-performing model was re-initialized with a fresh set of weights. Reusing weights from the previous iteration allowed the active learning procedure to build upon its previous choices, while re-initializing one model reduced the risk of the ensemble becoming stuck in local optima.

\subsection{Trial selection}
\label{sec:trial_selection}

Given an ensemble of psychological embeddings, the goal is to select trials that will maximize expected information gain. Given the enormous number of possible trials, it is not feasible to compute the information gain for all trials. Instead, the search space is reduced using a set of heuristics. First, we describe how information gain is computed for an arbitrary trial. Second, we describe the heuristics used to subsample a set of candidate trials.

The expected information gain of a candidate trial $\mathcal{T}_{m+1}$ is the difference between the Shannon entropy of the current embedding (using all observations collected so far) and the expected Shannon entropy if the embedding was inferred with the candidate trial
\begin{equation}
I\left(\rmZ, \ry| \mathcal{D}, \mathcal{T}_{m+1} \right) = H\left(\rmZ | \mathcal{D} \right) - H\left(\rmZ | \mathcal{D}, \mathcal{T}_{m+1}, \ry \right).
\end{equation}
The discrete random variable $\ry$ represents all possible outcomes associated with the candidate trial. Using the identity for mutual information and sample-based integral approximations, the expected information gain for a candidate trial can be computed using samples from the posterior \cite{Roads_Mozer_2019_BRM}. For a given candidate trial, the ensemble-level expected information gain is the mean expected information gain of all models.

The set of candidate trials is constrained using two heuristics that operate in tandem. The first heuristic influences which stimuli serve as a query stimulus. Query stimuli are stochastically sampled in proportion to stimulus entropy. Since stimulus entropy can be driven by aleatoric or epistemic uncertainty, the sampling probability is frequency adjusted to discourage perseverating on stimuli that have high aleatoric uncertainty,
\begin{equation}
    P\left(q\right) \propto \frac{1}{c_{q} + 1} \widetilde{H}(\mZ_{q} | \mathcal{D}),
\end{equation}
where $c_{q}$ indicates the number of times that an arbitrary stimulus $q$ has already served as a query. The tilde notation is used to indicate measures derived from an equal-weight ensemble average. For each iteration, 828 query stimuli are stochastically sampled without replacement.

The second heuristic influences which stimuli serve as a reference stimuli for the chosen queries. Given a query, references are stochastically sampled without replacement, in proportion to their ensemble-average expected similarity,
\begin{equation}
    P\left(r\right | q) \propto \E \tilde{s}\left(\mZ_{q}, \mZ_{r} \right).
\end{equation}
The set of eligible reference stimuli was restricted to the 500 nearest neighbor stimuli (1\% of the dataset), where nearest neighbors was also determined by ensemble-average, expected similarity. For each query, 10,000 candidate trials were assembled by choosing from the eligible set of reference stimuli. The eligible reference stimuli were limited to the 500 nearest neighbors in order to ensure that there was sufficient sampling of high similarity trials. Without this threshold, it is very unlikely that the sampled trials will adequately probe \emph{fine-grained} similarity structure. The ensemble-average information gain was computed for all candidate trials, and the top three trials for each query were retained (2484 trials). Three trials were retained in order to lessen the influence of any one participant, while also providing different information.

While the selection heuristics curb computational cost, they risk driving the active learning procedure into a local optimum. Given the large number of stimuli, any given stimulus is at risk of being embedded among highly dissimilar neighbors. Purely by chance, some stimuli may be repeatedly embedded among high dissimilar neighbors, preventing the detection of a signal that could move the stimulus to more similar neighbors. To offset this risk, 828 additional \emph{confirmation} trials were included in each iteration. A confirmation trial was assembled by first stochastically sampling a query in proportion to the ensemble-average entropy. Next, two references were randomly selected from among the 500 nearest neighbors and six references were randomly selected from stimuli outside of the 500-nearest neighborhood. If the stimulus was embedded in an appropriate neighborhood, participants will select the two references drawn from the 500 nearest neighbors. If not, participant choices will help move the stimulus to a more appropriate neighborhood.

\section{Evaluating dataset convergence}
\label{sec:convergence}

Since the human similarity judgments and companion embeddings are used to evaluate the internal representations of other models, it is critical that a sufficient amount of data has been collected. Convergence is primarily assessed by examining how the psychological embeddings change as more data is collected. We focus on three measures of the psychological embeddings: \emph{coarse-grained loss}, \emph{within-ensemble agreement}, and \emph{across-iteration agreement}. These measures and corresponding results are described in the remainder of this section.

\subsection{Coarse-grained loss}

A simple measure of convergence is to evaluate the ability of an ensemble to correctly predict a \emph{coarse-grained set} of similarity judgments. It is misleading to call this set of observations a test set. A basic requirement is that a test set be drawn from the same distribution as the training set. Since the coarse-grained trials are sampled randomly, they are unlikely to contain trials composed of highly similar images. As a consequence, this set will contain few observations that probe fine-grained similarity. We therefore refer to the corresponding loss (\ie, categorical cross-entropy) as coarse-grained loss.

By computing coarse-grained loss for ensembles trained at different iterations, we can see how generalization performance changes with additional data. Coarse-grained loss that has asymptoted at a minimum value provides some evidence that the psychological embeddings have converged.

\subsection{Within-ensemble agreement}

As an alternative to generalization performance, we can examine the learned similarity structure of the psychological embeddings \cite{Roads_Mozer_2019_BRM}. If a sufficient amount of data has been collected, then all models \emph{within} an ensemble should have comparable embeddings. The degree of agreement is quantified by comparing the implied pairwise similarity matrix of one model with the similarity matrix of another model. Since we have access to the posterior distribution, we can assembled the expected similarity matrix for each model. By computing the Pearson correlation between the upper diagonal elements of the similarity matrix, we can quantify the degree that the two models agree \cite{Shepard_1970}. Since there are three models, within-ensemble correlation is defined as the average over all pairwise model correlations.

\subsection{Consecutive-ensemble agreement}

Analogous to within-ensemble correlation, we can examine how the learned similarity structure differs \emph{across} iterations. If a sufficient amount of data has been collected, then including more data should not alter the inferred ensemble. To compute consecutive-ensemble correlations, the Pearson correlation is computed between the expected similarity matrix of ensembles belonging to consecutive iterations.

\subsection{Convergence results}

\begin{figure}[t]
\label{fig:evo}
\begin{center}
\includegraphics[width=1.\linewidth]{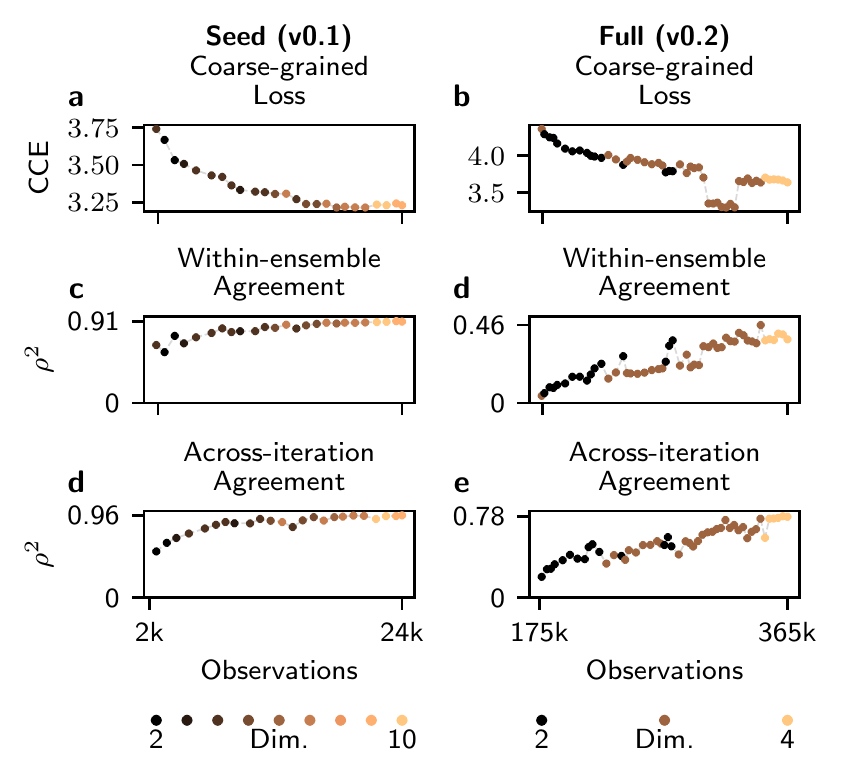}
\end{center}
\caption{Measures for assessing the convergence of ImageNet-HSJ. The left and right columns show measures for seed (v0.1) and full (v0.2) dataset respectively. Each panel plots the performance of a measure (ordinate) as the number of observations increases (abscissa). The first row \textbf{(a,b)}, demonstrates how categorical cross-entropy for a novel set of observations decreases as the amount of training data increases. It should be noted that the results in the two columns are not comparable since they use different coarse-grained sets. The second row \textbf{(c,d)}, shows that models within an ensemble become increasingly consistent as measured by the squared Pearson correlation between similarity matrices. Likewise, the third row shows that ensembles of consecutive iterations become increasingly stable as measured by the squared Pearson correlation between ensemble similarity matrices. Each point represents a metric evaluation of a single iteration (not all iterations are shown). Points are color-coded based on the selected dimensionality of the ensemble.}
\end{figure}

 All measures indicate that the psychological embeddings for the seed subset have converged, while the full dataset has not converged (Figure~\ref{fig:evo}). It should be noted that it is not appropriate to directly compare Figure~\ref{fig:evo}a and b. A different coarse-grained set is used for the seed and full dataset since since these two versions contain a different number of unique stimuli (1,000 vs 50,000 respectively). The dip in Figure~\ref{fig:evo}b clearly demonstrates how susceptible embeddings are to local optima.

\section{Using the dataset to evaluate target models}
\label{sec:evaluation}

Given a set of high-quality human similarity judgments, it is now possible to assess the human-likeness of representations from arbitrary models. While there are a variety of approaches for performing this assessment, we describe two straightforward metrics for evaluating a \emph{target model}. The first metric converts the 8-rank-2 trials into a set of triplet inequality relations and counts the proportion of triplets satisfied by the target model. The second metric examines the second order isomorphism between the internal representations of the target model and a psychological embedding \cite{Shepard_1970}.

\subsection{Triplet accuracy}

The first evaluation metric provides a minimalist approach for assessing the representations of a target model. This metric does not take advantage of the inferential machinery of a psychological embedding, but uses the similarity judgments directly. There are two primary advantages associated with this evaluation metric. First, this metric is less sensitive to issues of dataset convergence. While the psychological embeddings may change as more data is collected, the veracity of the observations does not change. Second, the researcher is only required to select a function for computing distances between representations of the target model.

Since each observed similarity judgment trial implies a set of triplet similarity relations \cite{Wah_2014}, we can convert an arbitrary observation into an equivalent set of triplet observations.  For example, consider an 8-rank-2 trial observation denoted as $q: a > b > [c, d, e, f, g, h]$, where $q$ indicates the query, $a$ indicates the first choice, $b$ indicates the second choice, and $c$-$h$ the remaining unselected references. This trial implies 13 triplet observations, \eg, $q: a > b$, $q: a > c$. 

The set of triplet observations is then used to compute the triplet accuracy for a target model. Given a triplet observation of the form $q: a > b$, and a distance function $d(\cdot,\cdot)$, we can determine if the target model correctly ranks a triplet,
\begin{equation}
    d\left(\mZ_{q}, \mZ_{a} \right) < d \left(\mZ_{q}, \mZ_{b} \right),
\end{equation}
where $\mZ$ indicates the stimulus representation of the target model. The mean number of correct predictions yields the triplet accuracy.

Given the nature of human similarity judgments, we do not expect a model to correctly rank all triplets. Different participants may not agree on perceived similarity and individuals may not be metric consistent \cite{Tversky1977:327PsychRev}.

\subsection{Psychological embedding correlation}

While triplet accuracy provides a simple method for assessing a target model, it is a relatively coarse metric that does not fully account for the similarity structure implied by the observations. Given sufficient observations, a stronger approach is to compare the internal representations of the target model to the internal representations of a psychological embedding. In order to execute this evaluation scheme, the researcher must make two choices. First, the research must select a psychological embedding. Second, the researcher much decide which similarity function to use when assembling a similarity matrix for the target model.

The psychological embedding correlation is computed in a similar manner as described previously. After assembling a pair-wise similarity matrix for the psychological embedding and the target model, we compute the Spearman correlation coefficient between the upper triangular portions of the two similarity matrices. Spearman correlation is used since we cannot assume the target model will yield similarities that are linearly related to similarities generated from the psychological embedding.

\subsection{Target model results}

The two evaluation metrics are demonstrated by assessing the representations for a handful of popular and successful target models. The selected supervised target models (and corresponding layer) are VGG16 (fc2), VGG19 (fc2), ResNet50 (avg\_pool) \cite{He_etal_2016}, ResNet101 (avg\_pool), ResNet152 (avg\_pool), ResNet50V2 (avg\_pool), Xception (avg\_pool) \cite{Chollet_2017}, DenseNet121 (avg\_pool) \cite{Huang_2017_CVPR}, InceptionV3 (avg\_pool) \cite{Szegedy_etal_2016_CVPR}, Inception-ResNetV2 (avg\_pool) \cite{Christian_etal_2017_inception}, MobileNet (avg\_pool) \cite{howard_etal_2017_mobilenets}. In addition to the supervised models, we include DeepCluster (VGG16, fc2) \cite{Caron_etal_2018}, since unsupervised models may be more capable of learning task-general representations. With the exception of DeepCluster, the pre-trained weights for the target models are obtained through Keras Applications \cite{Chollet_etal_2015_keras}. In addition to model-specific preprocessing, images are appropriately resized from the largest centered crop that preserves the aspect ratio. The weights and preprocessing procedure for DeepCluster are obtained from the corresponding GitHub repository.

Computing the triplet accuracy for the target models reveals that all models are better than chance (Table~\ref{tab:triplet}). Since computing triplet accuracy requires a choice of distance function, triplet accuracy is computed using L1, L2, and cosine distance. In all cases, cosine distance outperforms L1 and L2 distance. ResNet50 has the highest triplet accuracy for both the seed and full dataset.

For comparison, we also compute triplet accuracy for a psychological embedding. The psychological embeddings perform the best at predicting the triplet ratings, despite being blind to the stimuli. In contrast, the other target models can leverage approximately one million images worth of experience. While not a completely fair comparison since the psychological embedding is effectively trained on the triplet data, one could imagine performing a hold-one-trial-out test procedure, which would result in a very similar triplet accuracy.

\setlength\tabcolsep{5pt}.

\begin{table}
\label{tab:triplet}
\begin{center}
\begin{tabular}{|l|c|c|c|c|c|c|}
\hline
Target Model & \multicolumn{3}{|c|}{\bf{Seed (v0.1)}} & \multicolumn{3}{|c|}{\bf{Full (v0.2)}} \\
& L1 & L2 & cos & L1 & L2 & cos \\
\hline\hline
Psych. Emb. & -- & \bf{81.7} & -- & -- & \bf{80.7} & -- \\
\hline
VGG16 & 61.1 & 57.8 & \bf{69.6} & 67.4 & 63.1 & \bf{74.4} \\
VGG19 & 60.9 & 57.7 & \bf{69.8} & 67.3 & 63.1 & \bf{74.6} \\
ResNet50 & 62.3 & 62.5 & \bf{70.9} & 68.6 & 69.0 & \bf{75.2} \\
ResNet101 & 60.8 & 61.6 & \bf{70.2} & 68.0 & 68.8 & \bf{75.0} \\
ResNet152 & 60.7 & 61.3 & \bf{70.4} & 68.1 & 68.8 & \bf{75.2} \\
ResNet50V2 & 57.6 & 57.9 & \bf{67.0} & 65.4 & 66.7 & \bf{72.9} \\
Xception & 55.3 & 53.9 & \bf{56.3} & 66.2 & 65.1 & \bf{67.2} \\
DenseNet121 & 64.2 & 62.7 & \bf{67.3} & 70.9 & 70.0 & \bf{73.3} \\
InceptionV3 & 54.9 & 54.2 & \bf{55.6} & 65.1 & 64.7 & \bf{65.5} \\
Inception & & & & & & \\
\hspace{3mm}ResNetV2 & 53.2 & 53.0 & \bf{54.0} & 64.8 & 64.8 & \bf{65.2} \\
MobileNet & 64.7 & 64.3 & \bf{67.5} & 69.9 & 69.5 & \bf{72.8} \\
DeepCluster & & & & & & \\
\hspace{3mm}VGG16 & 56.8 & 58.0 & \bf{61.9} & 59.0 & 60.6 & \bf{65.8} \\
\hline
\end{tabular}
\end{center}
\caption{Triplet accuracy for various target models. Triplet accuracy measures the ability of a target model to correctly predict implicit triplet inequalities derived from the 8-rank-2 similarity judgments. The best performing result for each model is emphasized in bold. The high triplet accuracy of the psychological embedding demonstrates that there is substantial room for improvement.}
\end{table}

The psychological embedding correlations are computed by reusing the ensembles fit during the active learning procedure. Since computing embedding correlation requires a choice of similarity function, results are presented using a dot product and cosine similarity. Mirroring the triplet accuracy results, ResNet50 is the top performer. In the majority of cases, cosine similarity yields the best correlation for each target model. For comparison, if we compute the rank correlation between two psychological embeddings trained on the same data, the Spearman correlation between the corresponding expected similarity matrices is 0.81 and 0.62 for the seed and full dataset respectively.

\begin{table}
\label{tab:correlation}
\begin{center}
\begin{tabular}{|l|c|c|c|c|}
\hline
Target Model & \multicolumn{2}{|c|}{\bf{Seed (v0.1)}} & \multicolumn{2}{|c|}{\bf{Full (v0.2)}} \\
& dot & cos & dot & cos \\
\hline\hline
VGG16 & 0.25 & \bf{0.28} & 0.25 & \bf{0.27} \\
VGG19 & 0.27 & \bf{0.29} & 0.26 & \bf{0.27} \\
ResNet50 & 0.23 & \bf{0.36} & 0.25 & \bf{0.30} \\
ResNet101 & 0.18 & \bf{0.33} & 0.23 & \bf{0.29} \\
ResNet152 & 0.18 & \bf{0.34} & 0.24 & \bf{0.29} \\
ResNet50V2 & 0.26 & \bf{0.30} & 0.23 & \bf{0.24} \\
Xception & 0.01 & \bf{0.06} & \bf{0.06} & 0.04 \\
DenseNet121 & 0.17 & \bf{0.26} & 0.19 & \bf{0.22} \\
InceptionV3 & 0.01 & \bf{0.06} & \bf{0.04} & 0.03 \\
Inception & & & & \\
\hspace{3mm}ResNetV2 & \bf{0.00} & -0.00 & -0.00 & \bf{0.02} \\
MobileNet & 0.23 & \bf{0.29} & 0.19 & \bf{0.23} \\
DeepCluster & & & & \\
\hspace{3mm}VGG16 & 0.01 & \bf{0.11} & -0.02 & \bf{0.07} \\
\hline
\end{tabular}
\end{center}
\caption{The psychological embedding correlation for various target models. Embedding correlation computes the Spearman correlation between the similarity matrices of an ensemble of psychological embeddings and a target model.}
\end{table}

\section{Discussion}

Assessing how consistent a model's internal representations are with human judgments can offer a useful evaluation metric. Here, we presented a method to efficiently infer embedding spaces from human judgments for a large number of items. We were able to consider a problem an order of magnitude larger than previous work by leveraging variational inference, ensemble models, and active learning.

We demonstrated that human similarity judgments collected by our method are useful in evaluating model representations. Critically, these metrics can be used to assess models trained using either supervised or unsupervised procedures. We expected that representations from unsupervised models would be more general and conform better to human judgment, but, interestingly, DeepCluster was one of the worst performing models. Although we only evaluated one unsupervised model, this result highlights the challenging \emph{semantic gap} that any learning algorithm must bridge, which we hope Imagenet-HSJ will help close. 

Within the set of supervised models considered, the results reflected the history of the field. Initially, classification models that performed better on ImageNet also better corresponded with human behavioral and brain data \cite{Krizhevsky_etal_2012}. Although newer models continue to advance classification performance, we found that correspondence with human similarity judgments has not increased. An analogous trend can be observed in the model rankings on Brain-Score \cite{Schrimpf_etal_2020_brainscore}, which evaluates correspondences between brain activity and model representations.

Human similarity judgments are a rich source of information. They illuminate how humans structure the world and flexibly shift between organization schemes depending on context. We offered a method to infer such embedding spaces. Although we focused on using the embedding spaces to evaluate model representations, we expect there will be other potential applications for our method and dataset.

\subsubsection*{Acknowledgments}
NIH grant no. 1P01HD080679, Wellcome Trust Investigator Award no. WT106931MA and a Royal Society Wolfson Fellowship 183029 to B.C.L.


\end{document}